  \documentclass[journal,comsoc]{IEEEtran}
  \usepackage{graphicx}
  \usepackage[T1]{fontenc}
\ifCLASSINFOpdf
\else
\fi
\usepackage{amsmath}
\usepackage[cmintegrals]{newtxmath}
\usepackage{url,multirow,morefloats,floatflt,cancel,tfrupee}
\makeatletter

\AtBeginDocument{\@ifpackageloaded{textcomp}{}{\usepackage{textcomp}}}
\makeatother
\usepackage{colortbl}
\usepackage{xcolor}
\usepackage{pifont}
\usepackage[nointegrals]{wasysym}
\makeatletter

\def\mcWidth#1{\csname TY@F#1\endcsname+\tabcolsep}

\def\cAlignHack{\rightskip\@flushglue\leftskip\@flushglue\parindent\z@\parfillskip\z@skip}
\def\rAlignHack{\rightskip\z@skip\leftskip\@flushglue \parindent\z@\parfillskip\z@skip}

\usepackage{ifxetex}
\ifxetex\else\if@twocolumn\usepackage{dblfloatfix}\fi\fi

\AtBeginDocument{
\expandafter\ifx\csname eqalign\endcsname\relax
\def\eqalign#1{\null\vcenter{\def\\{\cr}\openup\jot\m@th
  \ialign{\strut$\displaystyle{##}$\hfil&$\displaystyle{{}##}$\hfil
      \crcr#1\crcr}}\,}
\fi
}

\AtBeginDocument{%
  \@ifpackageloaded{endfloat}%
   {\renewcommand\efloat@iwrite[1]{\immediate\expandafter\protected@write\csname efloat@post#1\endcsname{}}}{}%
}%
\usepackage{tabulary}

\let\lt=<
\let\gt=>
\def\processVert{\ifmmode|\else\textbar\fi}

\@ifundefined{subparagraph}{
\def\subparagraph{\@startsection{paragraph}{5}{2\parindent}{0ex plus 0.1ex minus 0.1ex}%
{0ex}{\normalfont\small\itshape}}%
}{}

\newcommand\role[1]{\unskip}
\newcommand\aucollab[1]{\unskip}
  
\@ifundefined{tsGraphicsScaleX}{\gdef\tsGraphicsScaleX{1}}{}
\@ifundefined{tsGraphicsScaleY}{\gdef\tsGraphicsScaleY{.9}}{}
\def\checkGraphicsWidth{\ifdim\Gin@nat@width>\linewidth
	\tsGraphicsScaleX\linewidth\else\Gin@nat@width\fi}

\def\checkGraphicsHeight{\ifdim\Gin@nat@height>.9\textheight
	\tsGraphicsScaleY\textheight\else\Gin@nat@height\fi}

\def\fixFloatSize#1{}
\let\ts@includegraphics\includegraphics

\def\inlinegraphic[#1]#2{{\edef\@tempa{#1}\edef\baseline@shift{\ifx\@tempa\@empty0\else#1\fi}\edef\tempZ{\the\numexpr(\numexpr(\baseline@shift*\f@size/100))}\protect\raisebox{\tempZ pt}{\ts@includegraphics{#2}}}}

\AtBeginDocument{\def\includegraphics{\@ifnextchar[{\ts@includegraphics}{\ts@includegraphics[width=\checkGraphicsWidth,height=\checkGraphicsHeight,keepaspectratio]}}}

\def\URL#1#2{\@ifundefined{href}{#2}{\href{#1}{#2}}}

\def\UrlOrds{\do\*\do\-\do\~\do\'\do\"\do\-}%
\g@addto@macro{\UrlBreaks}{\UrlOrds}

\@ifundefined{quoteAttrib}
	{}
	{}

\@ifundefined{titlequoteAttrib}
	{}{}

\newenvironment{title-quote}
	{\list{}{\fontsize{10pt}{12pt}\selectfont\leftmargin.5in\itshape\rightmargin\leftmargin}%
  \item\relax}
  {\endlist}

\makeatother

\makeatletter
\AtBeginDocument{\@ifpackageloaded{longtable}{%
\def\LT@makecaption#1#2#3{%
  \LT@mcol\LT@cols c{\hbox to\z@{\hss\parbox[t]\LTcapwidth{%
    \sbox\@tempboxa{#1{#2: } #3}%
    \ifdim\wd\@tempboxa>\hsize
      #1{#2: }\textsc{#3}%
    \else
      \hbox to\hsize{\hfil\box\@tempboxa\hfil}%
    \fi
    \endgraf\vskip\baselineskip}%
  \hss}}}
}{}}
\makeatother

   \makeatletter
  \def\fig@textbf{\textbf}
   \AtBeginDocument{\renewcommand\floatc@plain[2]{\setbox\@tempboxa\hbox{{\footnotesize#1.}\footnotesize\hskip.5em#2}%
    \ifdim\wd\@tempboxa>\hsize {\fig@textbf{\footnotesize#1.}}\footnotesize\hskip.5em#2\par
        \else\hbox to\hsize{\hfil\box\@tempboxa\hfil}\fi}}
    \makeatother
  
\usepackage{float}
\usepackage{booktabs}

\def\truncatedAt{1000}    
    \makeatletter
    \@ifpackageloaded{hyperref}{}{\usepackage[pdfborder={0 0 0},bookmarks=false]{hyperref}}
    \def\putUpgradeInfoBox{\@ifundefined{truncatedAt}{\def\truncatedAt{1000}}{}
    \def\up@width@one{\if@twocolumn .95\columnwidth\else .8\columnwidth\fi}%
    \def\up@width@two{\if@twocolumn .5\columnwidth\else .3\columnwidth\fi}%
    \vskip 2pc\nopagebreak
    \noindent\centering\fboxsep 7pt \fbox{\parbox{\up@width@one}
    {\vskip 1.7pc
    \begin{center}
    \fontfamily{phv}\fontsize{18pt}{18pt}\selectfont%
    \resizebox{\up@width@one}{!}{\parbox{\up@width@one}
    {\centering{IT'S TIME TO UPGRADE}}}\\[2pc]
    \fontsize{9.5pt}{13pt}\selectfont%
    \linespread{1}%
    \def\baselinestretch{1}\selectfont%
    \resizebox{\up@width@one}{!}{\parbox{\dimexpr(\up@width@one)}
    {\centering Dear user, you are able to see document only till the word limit
    of \truncatedAt~words. To view entire document upgrade your~plan.}}
    \vskip 2.5pc
    {\href{https://typ.st/2noUNUS}{\includegraphics[width=\up@width@two]{upgrade-box-click.pdf}}}
    \vskip 1.5pc
    \rule{\dimexpr(\up@width@one-6pt)}{.5pt}
    \par
    \vspace*{-1pt}\par
    \fontsize{10pt}{13pt}\selectfont
    \fontfamily{phv}\selectfont\href{https://typ.st/2xUBN7Q}{www.typeset.io}
    \end{center}
    }}}
    \makeatother
    
\begin{document}

	%
	
	\title{Region of Interest (ROI) based adaptive cross-layer system for real-time video streaming over Vehicular Ad-hoc NETworks (VANETs)}
	
	\author{Mohamed~Aymen~Labiod, Mohamed~Gharbi, Fran\c{c}ois-Xavier~Coudoux, Senior Member, IEEE, and Patrick~Corlay	
		\thanks{Mohamed Aymen~Labiod, Mohamed~Gharbi, Fran\c{c}ois-Xavier~Coudoux and Patrick~Corlay 
			are with Univ. Polytechnique Hauts-de-France, CNRS, Univ. Lille, YNCREA, Centrale Lille, UMR 8520 - IEMN, DOAE, F-59313 Valenciennes, France (e-mail: MohamedAymen.Labiod@uphf.fr).}}

	\maketitle 
	
	\begin{abstract}
		
Nowadays, real-time vehicle applications increasingly rely on video acquisition and processing to detect or even identify vehicles and obstacles in the driving environment. In this letter, we propose an algorithm that allows reinforcing these operations by improving end-to-end video transmission quality in a vehicular context. The proposed low complexity solution gives highest priority to the scene regions of interest (ROI) on which the perception of the driving environment is based on. This is done by applying an adaptive cross-layer mapping of the ROI visual data packets at the IEEE 802.11p MAC layer. Realistic VANET simulation results demonstrate that for HEVC compressed video communications, the proposed system offers PSNR gains up to 11dB on the ROI part.
		
	\end{abstract}

	
	%
	\IEEEpeerreviewmaketitle

	\section{ Introduction}
	
	With the development of vehicular communications, the need for efficient video transmissions with Quality of Service (QoS) guarantees is increasing. This is especially true for infrastructure-less applications requiring low latency such as Vehicle-to-Vehicle (V2V) systems for driver assistance \unskip~\cite{6170893}, but also for connected and autonomous vehicles. In emergency applications, visual information can be transmitted through Vehicle-to-Infrastructure (V2I) or Vehicle-to-Everything (V2X) communication patterns to help first responders to better prepare for victim assistance. Video delivery over vehicular networks can also be of great help for field agents in the case of video surveillance, law enforcement and traffic police applications \unskip~\cite{park_vehicular_2010}. However, in this kind of application, it is well known that the human viewer will not base his/her judgment on the entire scene but on some specific privileged regions of the scene relevant for his/her perception task. These so-called regions of interest (ROI) have a great influence on the perceived video quality and the subsequent visual analysis \unskip~\cite{2502676}. Indeed, they directly influence the ability and speed of human beings or autonomous systems to perceive, recognize and understand the surrounding world. Thereby, many video streaming applications use appropriate coding and bitrate control techniques that use the specific characteristics of video content\unskip~\cite{2502677}. Recently, many ROI-based encoders have emerged, developed in particular on the Advanced Video Coding (AVC) and High Efficiency Video Coding (HEVC) standards \unskip~\cite{fatani_robust_2012, meuel_region_2015}. However, due to severe channel transmission conditions, the Vehicular Adhoc NETwork (VANET) has a percentage of packets loss quite high and low reliability which does not promote video transmission. This can be all the more harmful if it affects the video ROI quality. Thus, an adequate end-to-end video transmission strategy and suitable standards are required.	
	
	In this letter, we propose a low-complexity adaptive cross-layer scheme that exploits the specific characteristics of video content, with the ROI, in the mapping of video packets at the IEEE 802.11p Medium Access Control (MAC) layer. Indeed, we intervene on two levels of the video transmission chain over vehicular network in the form of an adaptive cross-layer scheme. At the encoding level, after application of a low complexity ROI detection algorithm, the sequence is encoded with HEVC by granting better video quality to the detected ROI areas. At the MAC layer level, we perform an adjustment of the video mapping algorithm between the different Access Categories (AC). Thus, the proposed algorithm makes it possible to prioritize the transmission of the ROI video packets based on the network traffic load and the region area of each packet. On the receiver side, a stream reconstruction and error concealment algorithms are proposed for the corrupted video stream. Overall, realistic VANET simulation results demonstrate that for video communications, the proposed system offers significant end-to-end improvements in received video quality. Thereby, for the ROI part the average PSNR gains can exceed 11 dB compared to conventional video transmission schemes.
	
	The rest of the letter is organized as follows. In Section II we list some related works. In Section III we propose a description of the proposed solution. The simulation setting and the performances evaluations are respectively presented in Section IV and V. Conclusions are drawn in Section VI.

	\section{Related works }
	
	Video transmission improvement over VANET has been widely discussed in the literature. In particular, with the use of the WI-FI extension for vehicular networks, i.e. IEEE 802.11p. Moreover, IEEE 802.11p has variable QoS that supports differentiated service classes at the MAC layer \unskip~\cite{campolo_todays_2015, 259576:5809266}. Working at the source coder level, the authors in \unskip~\cite{garrido_abenza_source_2018} studied different coding options for the HEVC video encoder in order to improve the video quality perceived in VANETs networks. They used different intra-predictions refresh options with appropriate tile coding, a feature of the HEVC coder which divides an image into independent rectangular regions and their simulations were deployed in VANET urban scenarios. Belyaev et al. \unskip~\cite{6891339} introduced a video coding and transmission system for a VANET-enabled application with the use of IEEE 802.11p from a vehicle to a roadside unit. The developed system, dedicated to video surveillance for public transport security and road traffic control, has been validated experimentally. The system is based on a low complexity 3-D discrete wavelet transform applied to real-time automotive monitoring applications.
	
	Regarding ROI image coding, in several applications such as videoconferencing, video surveillance or telemedicine, video encoding taking into account the ROI has become an unavoidable subject. Indeed, users pay much more attention to the ROI areas, while less attention to the rest, called non-ROI. Therefore, maintaining the visual quality of the entire frame has become a major concern, especially in conditions of low bit rate \unskip~\cite{6200316,5167372}. Wu et al. \unskip~\cite{5337348} have proposed a mechanism for coding traffic surveillance videos by defining the areas covering vehicles as ROI in H.264 / AVC video compression, thereby preserving their quality. Their experimental results showed that the system works well in traffic surveillance videos.
	The authors in \unskip~\cite{pan_object-based_2019} proposed an object-based source coding method for video streaming over wireless networks to maintain constant video quality. Fatani et al.\unskip~\cite{fatani_robust_2012} have proposed a hierarchical video coding mechanism based on ROI scene separation for train-to-wayside wireless communication in driverless applications for the metro. The flow corresponds to the ROI area is transmitted through a reliable channel. This has ensured better transmission robustness and guarantees an acceptable level for the QoS perceived by the user. The creation of the different ROIs is based on the FMO (Flexible Macroblock Ordering) tool introduced in the new H.264 / AVC compression standard. Moreover, Meddeb et al.\unskip~\cite{meddeb_roi-based_2015} have considered an ROI-based rate control (RC) algorithm designed for HEVC. It has improved the quality of important regions considering independently coded regions within an ROI and helps to assess the ROI quality in poor channel conditions. The validation of the results was done on a flow transmitted over a lossy network modelized by a Gilbert-Elliott channel. 
		
	
	In the case of video transmission over IEEE 802.11e, improvements have also been made to the Enhanced Distributed Channel Access (EDCA). Indeed, Lin et al. \unskip~\cite{259576:5809274} have proposed an adaptive cross-mapping algorithm to improve MPEG-4 Part 2 video transmission over IEEE 802.11e wireless networks. Before them, Ksentini et al. \unskip~\cite{259576:5809276} were the pioneers with the idea of using the other Access Categories (ACs). They proposed a cross-layer static architecture based on an H.264 video stream. More recently, the authors \unskip~\cite{LABIOD201928,8585540} proposed a cross-layer mechanism based on the IEEE 802.11p standard to improve the packet mapping of HEVC video streaming in VANETs under low-delay constraints. They also proposed a low-complexity system based on robust encoding using Multiple Description Coding (MDC) in the video packets mapping at the IEEE 802.11p MAC layer \unskip~\cite{labiod_cross-layer_2019}.
	
	\section{Description of the proposed solution}

	In this section, we will briefly describe the technologies employed in the development of the different mechanisms associated to this work and then detail the proposed system.

	\subsection{The IEEE 802.11p standard}
	
	The Wireless Access in the Vehicular Environment (WAVE) is based on two standards categories: IEEE 802.11p for the PHY and MAC layers and IEEE 1609 for network management, security and other sides of VANETs \unskip~\cite{festag_standards_2015}. The spectrum band reserved for IEEE 802.11p is divided into seven 10 MHz channels, in the 5.850-5.925 GHz frequency range and is based on the Dedicated Short-Range Communication (DSRC) standard. It provides communication with a bit rate variation from 3 to 27 Mbps at theoretical distances up to 1000 meters. For the MAC layer, the IEEE 802.11p is based on the IEEE 802.11e standard with the use of EDCA for packet transmission. The EDCA provides sufficient QoS and is an improvement of Distributed Channel Access (DCA). However, instead of a single queue storing data frame, EDCA has four queues representing different levels of priority named AC. Each of these ACs is dedicated to a kind of traffic, namely Background (Bk or AC0), Best Effort (BE or AC1), Video (VI or AC2) and Voice (VO or AC3). Based on different Arbitration Inter-Frame Space Number (AIFSN) and Contention Window (CW) for the different ACs types. Thus, Voice traffic is given the highest priority while the Background the lowest one \unskip~\cite{259576:5809278,259576:5809273}.

	\subsection{Video Encoding}
	
	HEVC offers a gain in compression but also better errors resilience than its predecessors. Thus, the HEVC encoder improves video transmission in circumstances presenting severe transmission conditions such as low bandwidth networks with latency constraints \unskip~\cite{Psannis}. Moreover, four predictive structures types are proposed in the HEVC. These different structures can be envisaged depending on the targeted application. It depends on several parameters such as efficiency, computational complexity, time processing or error resilience techniques. In the All Intra (AI) structure, all the pictures are coded independently with only intra-predictions. Thus, this structure is most suitable for low latency constraints applications at the expense of a higher rate compared to the other three structures \unskip~\cite{259576:5809282}.

		\bgroup 
	\fixFloatSize{images/2ded5fbc-8c46-49ca-9913-252f3ef33f60-uproposed-system-letter-0.eps}
	\begin{figure*}[htbp!]
		\centering \makeatletter\IfFileExists{images/Proposedsystemletter.pdf}{\includegraphics[width=1.05\linewidth]{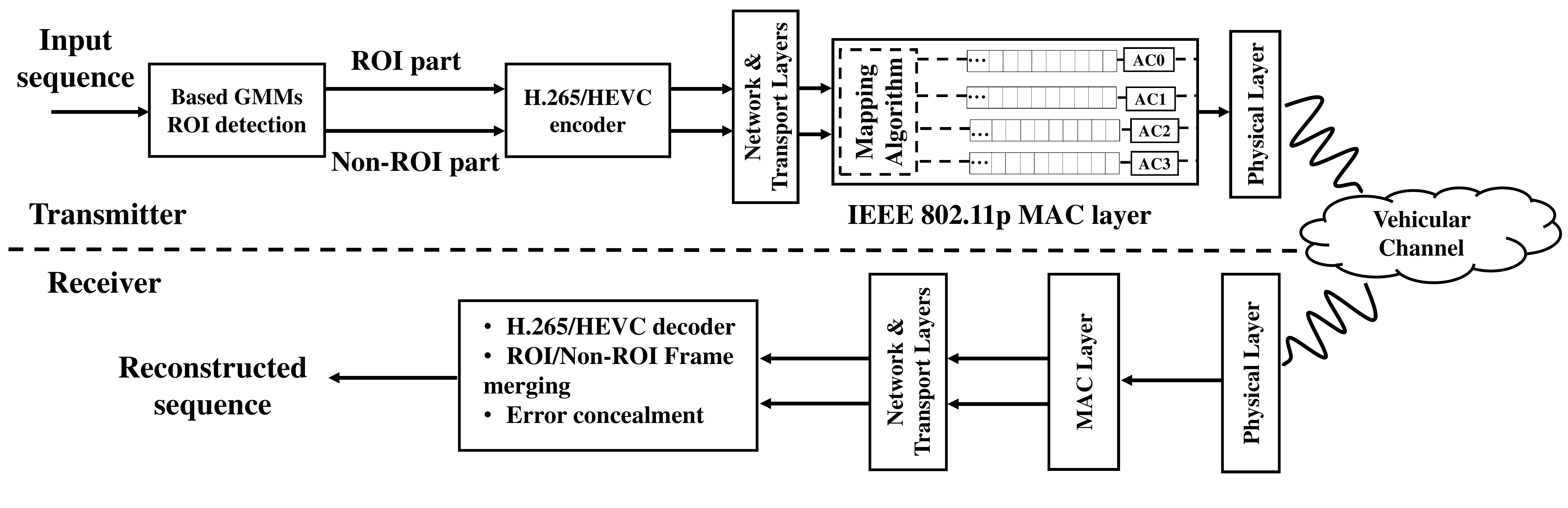}}{}
		\makeatother 
		\caption{{Representation of the adopted ROI based transmission scheme.}}
		\label{figure-1eb5a4b12034e1ad327e14a8f75ce28d}
	\end{figure*}
	\egroup
	
	Flexible Macroblock Ordering (FMO) is one of the characteristics of the H.264/AVC standard. It has been widely used to improve error resilience because each slice, i.e. a number of coded macroblocks, can be decoded independently of other slices. In the HEVC standard, a new feature called Tiles has been introduced that divides a picture into independent rectangular regions. Tiles are better suited for ROI coding than, e.g., slices. They ensure that temporal and spatial predictions within the ROI do not refer to pixels outside the ROI and that the ROI tiles are independently decodable from the non-ROI tiles \unskip~\cite{6547985}. In some works, tiles are combined with a tracking algorithm that can be used to dynamically adjust the ROIs size. However, tiles generally come along with bit rate overhead \unskip~\cite{sze_high_2014}.

	\subsection{Description of the proposed system}
	
	The proposed low-complexity cross-layer system exploits the specific characteristics of video content. The frames corresponding to the ROI flow are prioritized in the transmission in a VANET transmission scheme. Fig.~\ref{figure-1eb5a4b12034e1ad327e14a8f75ce28d} illustrates the adopted cross-layer design between the application layer and the IEEE 802.11p MAC layer.
	At the application layer, a real-time and low-complexity ROI detection algorithm is applied based on Gaussian Mixing Models (GMMs) \unskip~\cite{zhang_vehicles_2016}. Despite different ROI extraction methods exist, GMM is a classical parametric model due to its robustness and its efficiency in different scenes. Thereby, on the proposed system, the regions of interest are determined and are separated from the non-ROI regions. Then ROI and non-ROI are encoded in parallel with an AI H.265 / HEVC encoder. The encoding video quality of the ROI areas is held higher than that of non-ROI. At the encoder output, the generated stream is sent to the lower level of the protocol stack. At the receiver, if all the packets are correctly received, both regions can be correctly decoded and can be merged to reconstruct the video. The proposed system applies an error concealment mechanism in packet loss case. Thereby, the adopted frame copy mechanism has the advantage of being of low complexity with satisfactory results. Let’s recall the advantage of applying video processing discrimination. The objective is to protect the ROI flow at the expense of the non-ROI flow. To do this, we propose to use the other ACs in addition to the one used for the video, i.e. the \textit{AC}[2]. The other ACs used are \textit{AC}[1] dedicated for the best effort flow and \textit{AC}[0] for the traffic background, both being of lower priority. 
	
	The proposed adaptive mapping algorithm allocates dynamically for each video packet the most appropriated AC at the MAC layer. It takes into account the state of the network traffic load and the type of area of each frame packet, i.e. if ROI or non-ROI. The algorithm differentiates between the two regions areas, each region has a probability of mapping to lower priority ACs, defined as \textit{P}\ensuremath{_{\rm \_Region}}. It depends on the importance of the frame which means \textit{P}\ensuremath{_{\rm \_non-ROI}}{\textgreater}\textit{P}\ensuremath{_{\rm \_ROI}}.

	As previously mentioned, the mapping takes into consideration the state of the channel thanks to the AC queues filling state. Indeed, the more the MAC queue is filled, the more the network is overloaded. In order to deal with network congestion, two thresholds are established, \textit{threshold\_high} and \textit{threshold\_low}. Developed initially by Lin et al. in \unskip~\cite{259576:5809274} and based on the principle of the Random Early Detection (RED) mechanism the adaptive mapping algorithm is based on the following expression:
	
	\begin{equation}
	{P_{\_new}} = {P_{\_Region}} \times \frac{{qlen\left( {AC\left[ {2} \right]} \right) - qt{h_{low}}}}{{qt{h_{high}} - qt{h_{low}}}}\
	\label{moneq2}
	\end{equation}
	
	\setcounter{table}{1}
	\begin{table*}
		\centering
		\caption{{ Number of received packets for each mapping algorithm } }
		\label{table-wrap-4aa459235b694b52d30c4aeb536fcfb7}
		\begin{tabular}{|l|l|l|l|l|r|r|} 
			\hline
			Mapping Algorithm                                                     & \begin{tabular}[c]{@{}l@{}}Transmit \\ ROI Packets \end{tabular} & \begin{tabular}[c]{@{}l@{}}Received \\ ROI Packets \end{tabular} & \begin{tabular}[c]{@{}l@{}}Transmit \\non-ROI Packets \end{tabular} & \begin{tabular}[c]{@{}l@{}}Received \\non-ROI Packets \end{tabular} & \multicolumn{1}{l|}{\begin{tabular}[c]{@{}l@{}}Transmit \\Packets \end{tabular}} & \multicolumn{1}{l|}{\begin{tabular}[c]{@{}l@{}}Received \\Packets \end{tabular}}  \\ 
			\hline
			EDCA                                                                  & \multicolumn{1}{c|}{/}                                         & \multicolumn{1}{c|}{/}                                                              & \multicolumn{1}{c|}{/}                                                                & \multicolumn{1}{c|}{/}                                                               & 3900                                                                            & 2533                                                                             \\ 
			\hline
			Algorithm in \cite{LABIOD201928}                                                     & \multicolumn{1}{c|}{/}                                                              & \multicolumn{1}{c|}{/}                                                            & \multicolumn{1}{c|}{/}                                                                & \multicolumn{1}{c|}{/}                                                               & 3900                                                                            & 3721                                                                             \\ 
			\hline
			\begin{tabular}[c]{@{}l@{}}Based ROI \\adaptive mapping \end{tabular} & \multicolumn{1}{r|}{1201}                                      & \multicolumn{1}{r|}{1196}                                      & \multicolumn{1}{r|}{2993}                                         & \multicolumn{1}{r|}{2871}                                         & 4194                                                                            & 4067                                                                             \\
			\hline
		\end{tabular}
	\end{table*}
	
	\setcounter{table}{2}
	\begin{table*}
		\caption{{ Average PSNR for each mapping algorithm } }
		\label{table-wrap-2}
		\centering
		\begin{tabular}{|c|c|c|c|c|c|c|} 
			\toprule
			\multicolumn{1}{|l|}{}                                               & \multicolumn{6}{c|}{PSNR (dB)}                                                                                                                                                                                                                                                                                                                                                              \\ 
			\hline
			\begin{tabular}[c]{@{}c@{}}Mapping \\Algorithm\end{tabular}          & \begin{tabular}[c]{@{}c@{}}Encoding \\ROI \end{tabular} & \begin{tabular}[c]{@{}c@{}}Received \\ROI \end{tabular} & \begin{tabular}[c]{@{}c@{}}Encoding \\non-ROI \end{tabular} & \begin{tabular}[c]{@{}c@{}}Received \\non-ROI \end{tabular} & \begin{tabular}[c]{@{}c@{}}Encoding \end{tabular} & \begin{tabular}[c]{@{}c@{}}Received \end{tabular}  \\ 
			\hline
			EDCA                                                                 & 34.95                                                       & 23.48                                                           & 31.28                                                          & 28.35                                                             & 31.07                                                   & 27.02                                                        \\ 
			\hline
			\begin{tabular}[c]{@{}c@{}}Algorithm in  \\ \cite{LABIOD201928}\end{tabular}           & 34.95                                                       & 24.85                                                           & 31.28                                                          & \textbf{30.03 }                                                            & 31.07                                                   & 28.54                                                        \\ 
			\hline
			\begin{tabular}[c]{@{}c@{}}Based ROI \\adaptive mapping\end{tabular} & 35.30                                                       & \textbf{34.93 }                                                          & 30.40                                                          & 27.91                                                             & 30.23                                                   & \textbf{29.07  }                                                      \\
			\bottomrule
		\end{tabular}
	\end{table*}

	Where \textit{P}\ensuremath{_{\rm \_Region}} is the initial probability of each region, \textit{qlen}(AC[\textit{2}]) is the actual state of the video queue length, \textit{qth}\ensuremath{_{\rm high}} and \textit{qth}\ensuremath{_{\rm low}} are thresholds that define the manner and degree of mapping to lower priority ACs.
	
	The algorithm behaves as follows: when \textit{qlen}(AC[\textit{2}]) is less than qth\ensuremath{_{\rm low}}, all packets are mapped in \textit{AC}[2]. When \textit{qlen}(AC[\textit{2}]) is between \textit{qth}\ensuremath{_{\rm low}} and \textit{qth}\ensuremath{_{\rm high}}, \textit{P}\ensuremath{_{\rm \_new}} sets the packet probability mapped to \textit{AC}[1]. And finally, when \textit{qlen}(AC[\textit{2}]) is greater than \textit{qth}\ensuremath{_{\rm high}}, video packets are mapped in \textit{AC}[1] with \textit{P}\ensuremath{_{\rm \_new}} probability of being mapped to \textit{AC}[0].	
	
	\bgroup
\fixFloatSize{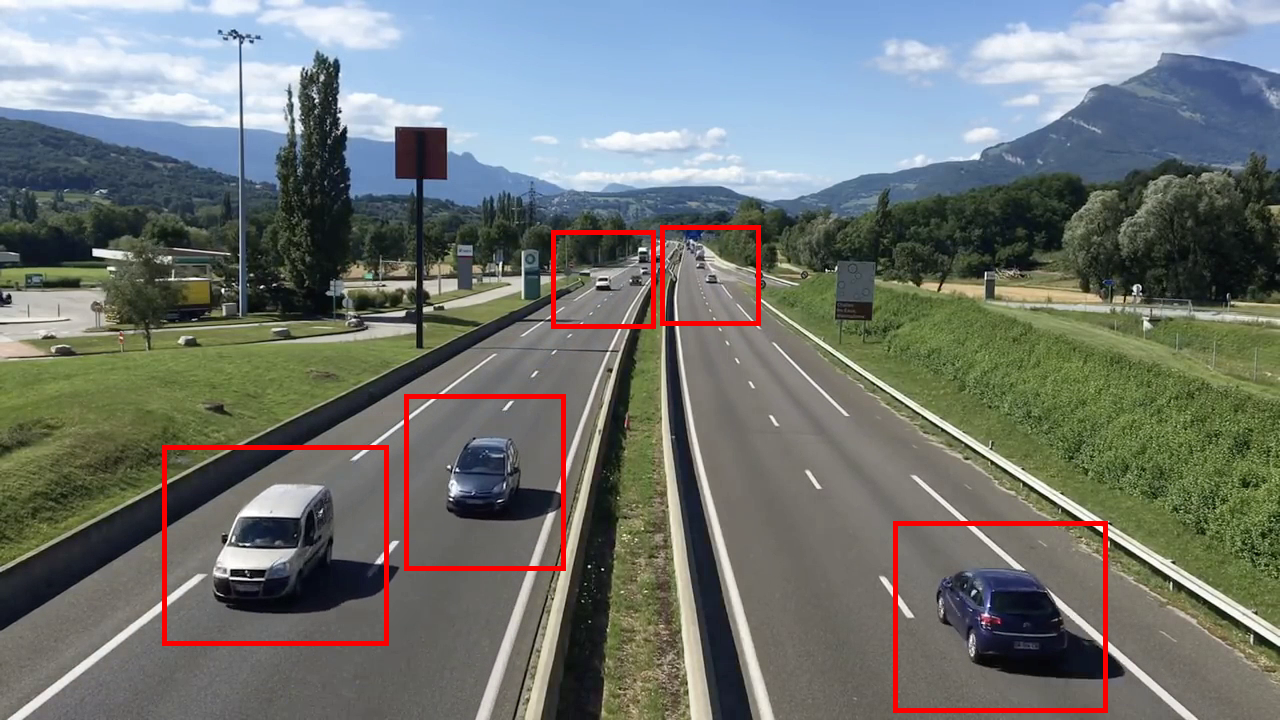}
\begin{figure}[!htbp]
	\centering \makeatletter\IfFileExists{images/2-zone_ROI_500.png}{\includegraphics[width=1.0\linewidth]{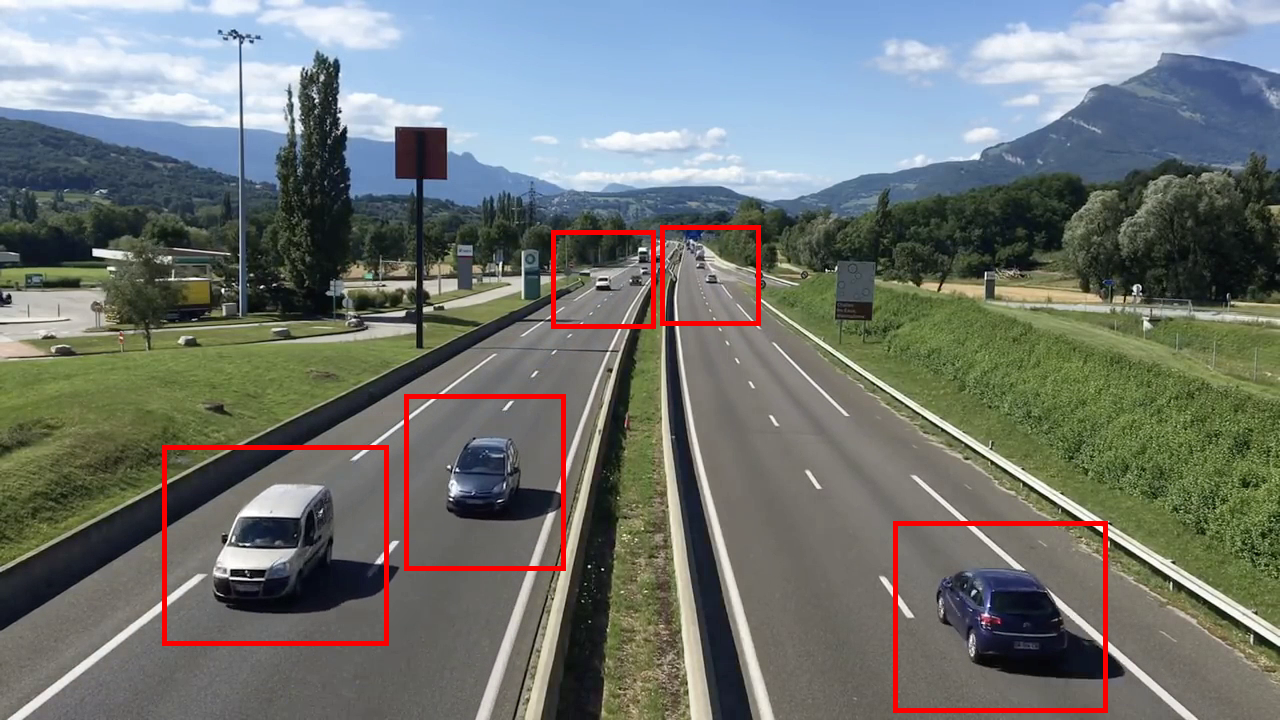}}{}
	\makeatother 
	\caption{{Illustration of a ROI based fixed camera on a European highway.}}
	\label{figure-5650f277e75d8c94f5f42e6e1e8331c4}
\end{figure}
\egroup

	\section{Simulation and parameters setting }

We have modeled a framework allowing a realistic simulation of video transmission in a vehicular environment. It consists of a vehicular traffic simulator, a network simulator and a video encoder / decoder. Thereby, we used Simulation of Urban Mobility (SUMO) \unskip~\cite{259576:5809280} to model the behavior of vehicles from urban traffic maps imported from OpenStreetMap (OSM) of Valenciennes (France) \unskip~\cite{259576:5818119}. Moreover, SUMO is an open source simulator, it models the behavior of vehicles with urban mobility and takes into account the interaction of vehicles with each other, traffic signals, junctions, etc. We also used NS2 as a network simulator to which we integrated the Evalvid tool \unskip~\cite{259576:5809281}. This tool allows us to transmit the video and reconstruct it on the receiver side. We also used the HM reference model (16.16) for video encoding \unskip~\cite{259576:5818121}.

\setcounter{table}{0}
\begin{table}[!htbp]
	\caption{{Simulation parameters of the VANET scenario and the measured metrics} }
	\label{table-wrap-7afe70dec01a7c70bf76894f5c7a3b32}
	\def\arraystretch{1}
	\ignorespaces 
	\centering 
	\begin{tabulary}{\linewidth}{LL}
		\hline 
		Parameters  & Value\\
		\hline 
		Number of vehicles    &
		100\\
		Radio-propagation model   &
		TwoRayGround\\
		Maximum Transfer Unit (MTU) &
		1024\\
		Routing protocol &
		Ad-hoc On-Demand Distance Vector (AODV)\\
		Transport protocol &
		User Datagram Protocol (UDP)\\
		Measured metrics   &
		Peak Signal to Noise Ratio (PSNR)\mbox{}\protect\newline Packet delivery ratio\\
		\hline 
	\end{tabulary}\par 
\end{table}
	
For the choice of the test sequence, we choose a sequence of a fixed camera of an overtaking on a European highway as shown in Fig.~\ref{figure-5650f277e75d8c94f5f42e6e1e8331c4}. The sequence lasts 10 seconds with an image resolution of $1280\times720$ pixels and a frame rate of 30 fps. The radio propagation model used is the TwoRayGround allowing to give a fairly realistic representation of the vehicular channel, the standard used is the IEEE 802.11p. As for the transport layer protocol and in order to guarantee a minimum latency, we have chosen to work with User Datagram Protocol (UDP). Table~\ref{table-wrap-7afe70dec01a7c70bf76894f5c7a3b32} summarizes the main parameters of the simulation and the measured metrics. The used system parameters are as follows: the value of \textit{P}\ensuremath{_{\rm \_Region}}, is fixed at 0 for the ROI and 0.8 for the non-ROI.
	
	\section{Performance evaluation }

To demonstrate the effectiveness of the proposed system, several experiments were carried out to evaluate three different mapping methods: first, the EDCA algorithm, the conventional video transmission scheme, implemented in the current IEEE 802.11p vehicular standard. Then, the enhanced adaptive cross-layer algorithm \cite{LABIOD201928} and finally our current system. For the first two algorithms, in the video encoding, no distinction is applied regarding the characteristics of video content. Thus, the video is encoded at 3 Mbps. In the adopted simulations, other types of stream coexist with video transmission in order to guarantee a realistic simulation. Indeed, we simulate voice traffic in the \textit{AC}[3] but also a TCP flow in the \textit{AC}[1] and a UDP flow on the \textit{AC}[0]. For our current system, ROI raw video is encoded with better video quality than the non-ROI part but keeping the entire sequence bit rate equivalent to that of other methods.

The simulations results obtained by the enhanced adaptive cross-layer \cite{LABIOD201928} and the EDCA algorithms are illustrated Table ~\ref{table-wrap-4aa459235b694b52d30c4aeb536fcfb7} in terms of correctly received packets. The performance gain provided by the enhanced solution \cite{LABIOD201928} is explained by the use of the two-lower priority ACs, i.e. \textit{AC}[1] and \textit{AC}[2]. While Table ~\ref{table-wrap-2} shows the PSNR average gain provided with a 1.5 dB gain. However, we note that for the ROI part, the two algorithms have a low PSNR values which is explained by the more pronounced movement in this part of the sequence. Hence, when a packet is lost, this generates an image freeze which makes a big difference compared to the original video. Regarding the proposed ROI based system, we note a greater packets number of the overall sequence, i.e. 4194 packets. This can be explained by the ROI coding and the video packets segmentation. Whereas, the ROI based algorithm allows a better reception of packets corresponding to the ROI part which results in better video quality perceived by the user at reception.
Regarding the PSNR values, the proposed algorithm has more than 11 dB gain compared to the EDCA for the ROI part and 10 dB gain compared to the algorithm of \cite{LABIOD201928}. The entire sequence received from the proposed system also exhibits better PSNR than the other two methods. Visually, we confirm that the quality of the received sequences with the proposed ROI-based adaptive cross-layer system fluctuates much less. This is particularly true for the ROI areas, which clearly translates into a better quality of experience for the final viewer.

	\section{Conclusion}
	
The letter proposes a cross-layer system, using the video ROI, allowing a video transmission improvement in low latency application over vehicular networks. The proposed system allows cross-layer packet classification based on the IEEE 802.11p protocol. Indeed, the proposed strategy is based on the video packets mapping in the most appropriate ACs in order to offer better QoS. The proposed solution uses video information at the MAC layer in a cross-layer scheme. Thus, the video ROI information and the MAC AC buffer filling allow the proposed algorithm to choose the best option for video mapping. The results established in a realistic vehicle environment illustrates a QoS improvement and an end-to-end video quality enhancement.


	
	%

	\bibliographystyle{IEEEtran}
	
	\bibliography{article}
	
\end{document}